\documentclass[10pt,twocolumn,letterpaper]{article}

\usepackage{cvpr}      

\usepackage[dvipsnames]{xcolor}

\definecolor{cvprblue}{rgb}{0.21,0.49,0.74}
\usepackage[pagebackref,breaklinks,colorlinks,citecolor=cvprblue]{hyperref}

\def\paperID{*****} 
\def\confName{CVPR}
\def\confYear{2024}

\title{Technical Report for CVPR 2024 WeatherProof Dataset Challenge: Semantic Segmentation on Paired Real Data}

\author{\quad Guojin Cao$^*$\\
{\tt\small    \quad 1244968517@qq.com}
\and
\quad Jiaxu Li$^*$\\
{\tt\small \quad 15319754952@163.com}
\and
\quad Jia He$^*$\\
{\tt\small \quad hejia20102@outlook.com}
\and
\quad Ying Min\\
{\tt\small \quad myingxd@163.com}
\and
\quad Yunhao Zhang\\
{\tt\small \quad yunhao.xa@qq.com}
}

\begin{document}
\maketitle
\def\thefootnote{*}\footnotetext{These authors contributed equally to this work}\def\thefootnote{\arabic{footnote}}

\begin{abstract}
This technical report presents the implementation details of 2nd winning for
CVPR'24 $UG^2$+ WeatherProof Dataset Challenge.
This challenge aims at semantic segmentation of images degraded by various
degrees of weather from all around the world.
We addressed this problem by introducing a pre-trained large-scale vision
foundation model: InternImage, and trained it using images with different levels of noise.
Besides, we did not use additional datasets in the training procedure and utilized dense-CRF as post-processing in the final testing procedure.
As a result, we achieved 2nd place in the challenge with 45.1 mIOU
and fewer submissions than the other winners.
\end{abstract}    
\section{Method}
\subsection{Baseline}
We utilized InternImage-H~\cite{wang2023internimage} of which detector is
Mask2Former in our experiment, serving as our baseline. Considering the unique characteristics of the WeatherProof dataset, we first compared the results of training on clean images and testing on degraded images versus both training and testing on degraded images. Differing from the conclusion in~\cite{gella2023weatherproof}, we found that both strategies yielded nearly mIOU values. 
We believe that the disappearance of performance differences stems from
InternImage's training approach for experimentation. Additionally, we also tried InternImage-XL, of which performance was significant inferior to InternImage-H, and ultimately abandoned it.

\subsection{Additional Denoised Data}
To facilitate the acquisition of additional visual information and robust representations during training, we incorporate clean images, noisy images, and denoised images derived from a denoising network into our selected model for joint learning.
Specifically, we employ DA-Clip~\cite{luo2023controlling} as our denoising
network to perform dehazing, deraining, and desnowing tasks on each input
degraded image, thereby enabling the model to leverage these diverse datasets and refine its understanding of visual patterns.
This component aims to enable the model to utilize these images with varying
levels of noise (clean, partially denoised, and noisy) to increase generalization ability and prevent overfitting.
It is worth noting that DA-Clip is employed only for data augmentation purposes and we do not integrate the network into our inference procedure.

\subsection{Postprocessing}
We used the model trained with different data combinations among clean, degraded, and denoised images to predict the result.
We observed significant fluctuation in accuracy across various models for distinct categories, highlighting the need for category-specific optimization and evaluation strategies.
To leverage the strengths of individual models, we integrated the prediction results of different versions through a voting strategy. 
As a result, mlOU has significantly improved as shown in~\cref{tab:per}.

We utilized dense-CRF~\cite{krahenbuhl2011efficient} as the post-processing
method.
Dense-CRF can combine the relationship between all pixels in the original image and process the segmentation results obtained by our trained model
model, optimize the rough and uncertain labels in the classified image, correct the fragmented misclassified areas, and obtain more detailed and smoother segmentation boundaries.
As shown in~\cref{tab:per}, the dense connection at the pixel level greatly improves the accuracy of segmentation and labeling while results after dense-CRF are more accurate.
Moreover, we utilized morphological transformations to eliminate isolated points in the segmentation results, enhancing edge connectivity, shown in ~\cref{tab:per}.

\begin{table*}[t!]
\centering
\begin{tabular}{c|c|c|c|c|c|c|c|c|c|c}
\toprule
\textbf{Model}& \textbf{mIOU} & building & structure & road & sky & stone & t.-grass & t.-other & t.-snow & tree \\
\hline
InternImageH & 39.61 & 53.16 & 25.44 & 25.13 & 80.62 & 14.10 & 47.69 &
50.71 & 38.66 & 60.61 \\
\hline
+E & 43.60 & 54.40 & 34.29 & 27.42 & 80.09 & 15.63 & 53.53 & 51.70 &
51.69 & 67.29 \\
\hline
+E+D & 44.48 & 56.86 & 36.24 & 28.33 & 81.23 & 19.45 & 53.06 & 52.33 &
51.39 & 65.92 \\
\hline
+E+D+M & 45.10 & 56.66 & 36.73 & 29.30 & 81.65 & 20.60 & 52.83 & 52.54 &
54.07 & 67.83 \\
\bottomrule
\end{tabular}
\caption{The performance of some notable uploads, here E represents model ensemble, D represents dense-CRF and M indicates morphological transformation}
\label{tab:per}
\end{table*}

\begin{figure*}
    \centering
    \includegraphics[width=1.9\columnwidth]{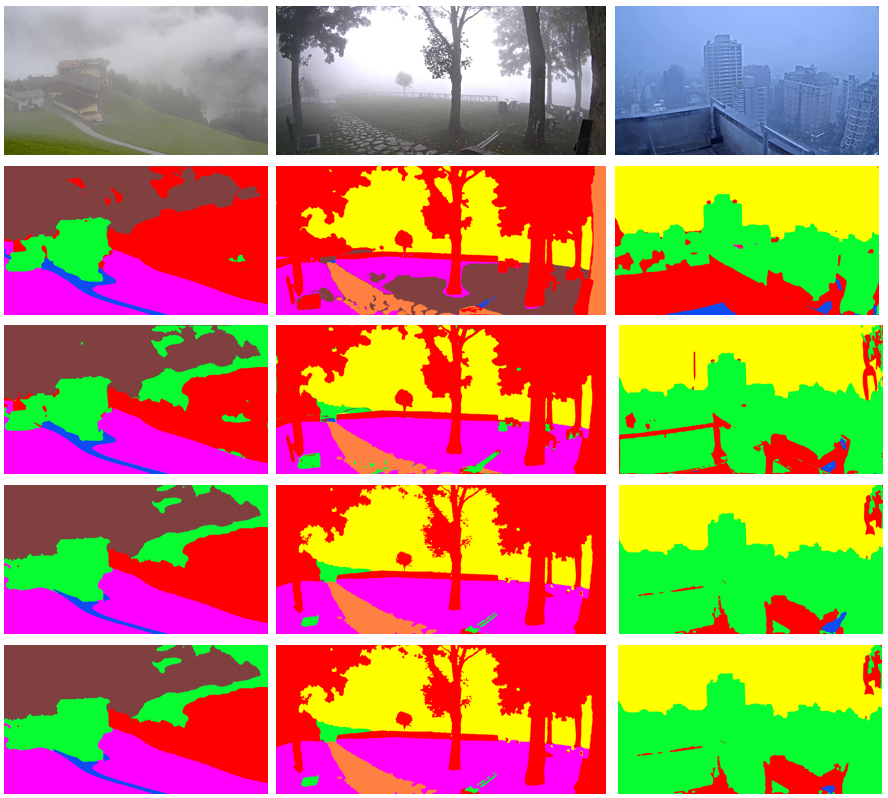}
    \caption{Visualization of segmentation results, from top to bottom: raw image, baseline+model ensemble, baseline+model ensemble+dense-CRF, and baseline+model ensemble+dense-CRF+morphological transformation (final submitted).}
    \label{fig:seg_res}
\end{figure*}

\section{Experiment Setup}
We trained our model on 2 Nvidia L40S GPUs with batch sizes 2. 
Each model is trained with 40000 iterations with an initial warm-up learning rate $10^{-5}$ and follows poly policy with power 0.9 and minimized learning rate $10^{-4}$. The optimizer is AdamW with betas ($0.9$, $0.999$) and weight decay $0.05$.
The training is conducted on Pytorch1.13.0 + cuda12.1.

For the training dataset, we did not use any additional datasets. In the final test
phase, for evaluation fair, we did not set the validation set as a part of the training set.
\cref{tab:per} demonstrates the segmentation results of some notable uploads and~\cref{fig:seg_res} shows some visualization examples.
{
    \small
    \bibliographystyle{ieeenat_fullname}
    \bibliography{main}
}


\end{document}